\documentclass[11pt]{article}
\usepackage{konvens2019}
\usepackage{mathptmx}
\usepackage[scaled=.90]{helvet}
\usepackage{courier}

\usepackage{url}
\usepackage{latexsym}

\usepackage{subcaption}

\makeatletter
\newcommand{\@BIBLABEL}{\@emptybiblabel}
\newcommand{\@emptybiblabel}[1]{}
\makeatother
\usepackage[hidelinks]{hyperref}

\usepackage{amsmath}
\usepackage{amssymb}
\usepackage{booktabs}
\usepackage{graphicx}

\usepackage{xcolor}
\definecolor{babypink}{rgb}{0.96, 0.76, 0.76}
\definecolor{amber}{rgb}{1.0, 0.75, 0.0}
\definecolor{burgundy}{rgb}{0.5, 0.0, 0.13}
\definecolor{cardinal}{rgb}{0.77, 0.12, 0.23}
\definecolor{carnelian}{rgb}{0.7, 0.11, 0.11}
\definecolor{blush}{rgb}{0.87, 0.36, 0.51}
\definecolor{cornellred}{rgb}{0.7, 0.11, 0.11}

\title{Argumentative Relation Classification as Plausibility Ranking}

\author{Juri Opitz \\
 Leibniz ScienceCampus ``Empirical Linguistics and Computational Language Modeling''\\
  Department for Computational Linguistics \\
  Heidelberg University \\
  69120 Heidelberg \\
  {\tt opitz@cl.uni-heidelberg.de} 
  \\}

\date{}

\begin{document}
\maketitle
\begin{abstract}
We formulate argumentative relation classification (support vs.\ attack) as a text-plausibility ranking task. To this aim, we propose a simple reconstruction trick which enables us to build minimal pairs of plausible and implausible texts by simulating natural contexts in which two argumentative units are likely or unlikely to appear. We show that this method is competitive with previous work albeit it is considerably simpler. In a recently introduced content-based version of the task, where contextual discourse clues are hidden, the approach offers a performance increase of more than 10\% macro F1. With respect to the scarce \textit{attack}-class, the method achieves a large increase in precision while the incurred loss in recall is small or even nonexistent. 

\end{abstract}

\section{Introduction}
\label{sec:intro}
 
 \textit{Argumentative relation classification} (ARC) is dedicated to determining the class of the relation which may hold between two arguments or elementary argumentative units, EAUs\footnote{Here, we use the term \textit{elementary argumentative units} to denote clauses or small clause-complexes -- e.g., \textbf{(0)}, \textbf{(1)} or \textbf{(2)}) -- which can be `instantiated' in an argumentative debate.}. For instance, consider the following \textit{premises} given the \textit{topic} or \textit{conclusion} \textbf{(0)} \textit{``Overall, marijuana is detrimental to your health.''}:

\begin{description}
    \item[(1)] \textit{Use of marijuana causes chronic bronchitis and airflow obstruction.}
    \item[(2)] \textit{Cannabis does not need to be smoked to receive its potential health benefits.}
    \label{ex:0}
\end{description}

\begin{figure}
    \centering
    \includegraphics[scale=0.67]{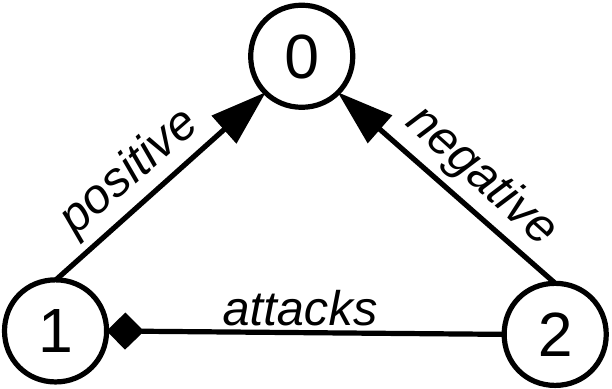}
    \caption{A small argumentation graph containing two general types of relations: premise-topic relations (class: negative/positive) and premise-premise relations (class: supports/attacks).}
    \label{fig:smallgraph}
\end{figure}
In this case, \textbf{(1)} has a \textit{positive stance} towards the conclusion \textbf{(0)}; in contrast to \textbf{(2)}, which has a negative stance towards the conclusion. Additionally, but not less importantly, we can say that \textbf{(2)} weakens \textbf{(1)} -- it casts doubt about its generality by hinting at cannabis application methods which do not involve combustion or inhalation. In this work, we summarize all relations which aim at undermining or weakening another argument or premise (`undercut', `rebuttal', etc.) as \textit{attack}.\footnote{For a more `in-depth' view and discussion of argumentative relations we refer the reader to, e.g., \newcite{Pollock:1995:CCB:526901}, \newcite{wd2009} and \newcite{besnard2014constructing}.} The EAUs from our example and their connecting relations are outlined in the graph in Figure \ref{fig:smallgraph}. 

In a rhetorically structured argumentative text\footnote{E.g., an argumentative essay.}, \textbf{(1)} and \textbf{(2)} may appear in configurations such as \textit{On the one hand \textbf{(1)}, on the other \textbf{(2)}}; \textit{\textbf{(1)}, however, \textbf{(2)}}, etc. Under these circumstances, discourse \textit{context} can predict argumentative relations very well. However, when moving from such `closed scenario' to a more `open-world setting', e.g., where EAUs have been mined from heterogeneous documents, we need to determine relations based on their \textit{content}. In this paper, we show that our method works well in both scenarios. In fact, it is in the more general 
and more difficult content-based setting, where our method provides the most benefits over previous work.

Systems which have learned to predict general argumentative relations have a decisive advantage when compared to systems that have `only' learned to predict argumentative stances: in an argumentative debate, often a debater does not choose to bring forth any argument which supports their stance on the topic. Instead, or additionally, they may choose to select an argument which also attacks the opponent's most recent argument. Therefore, we need not only knowledge about the stances of arguments towards topics, but also about relations to other arguments. 
Our experiments show that our approach is a step towards this goal.

The remainder of this paper is structured as follows: After discussing related work in Section \ref{sec:relw}, we propose a simple reconstruction trick which allows us to embed an argumentative source-target pair in a relational discourse context yielding a plausible and implausible text variant (Section \ref{sec:rec}). 
In Section \ref{sec:ex}, we conduct experiments and ablation studies using (i) a standard task setup, where systems are allowed to see EAUs in their document context and (ii) a more difficult `content-focused' task setup where systems are only allowed to see the spans of the EAU clauses. \label{fn:code}The code for this paper is available at \url{https://gitlab.cl.uni-heidelberg.de/opitz/pr4arc}

\section{Related work}
\label{sec:relw}
In this section, we first provide an overview of the data, and the data issues people are confronted with when developing argumentative relation classification (ARC) systems. We proceed with an overview of existing ARC approaches and conclude by touching on other related tasks. 

\paragraph{Argumentative relation data} For general argumentative relations, not many data sets have been developed. One of the largest data sets consists of 402 argumentative student essays and is henceforth denoted by \textsc{Essay} \cite{stab2014identifying,stab2017parsing}. It has been annotated, i.a., with EAU clauses and more than 3,000 relations which hold among them. By \textsc{Essay-Content}, we denote a version of \textsc{Essay} from which discourse context is stripped and systems can only access the spans of EAU clauses \cite{opitzdissec}. This setup is more difficult since systems have to learn to model the \textit{content} of two EAUs in order to successfully predict their relation. 
\textsc{Essay} and \textsc{Essay-Content} will be more extensively described in Section \ref{par:data}, where we also show that our method is efficient across both setups.

Another data set which is annotated with in-depth argumentative annotations is the Microtext corpus covering a variety of political debates in Germany \cite{PeldszusStede-ECA:16}. While it has been annotated with a more fine-grained set of relations (e.g., \textit{rebutting attack, undercutting attack,  linked  support, example support}) it is rather small in size (the recently extended version \cite{skeppstedt-etal-2018-less} contains about 700 relation tuples). Similar to \textsc{Essay-Content}, a variant of the Microtext corpus exists where argumentative units are detached from discourse context \cite{wachsmuth-etal-2018-argumentation}. We believe that systems that have learned to predict argumentative relations based on the \textit{content} of argumentative units have  advantages over systems which focus too much on contextual discourse clues. For example, content-focused systems can better be expected to solve large-scale cross-document tasks where EAUs are mined from many heterogeneous documents. 
Our reconstruction trick provides one step towards this goal: it exploits \textit{potential} discourse configurations without depending on seeing the \textit{true} discourse context.

A key reason for the data scarcity of annotated general argumentative relations is that creating high-quality data for `premise-premise' relations is a challenging task. Perhaps, it is more challenging than creating data for argumentative stance detection since topics or conclusions are often `a-priori' well understood (e.g., \textit{Cannabis should be legalized}) and always occur as the stance-relation target. In that sense, it may be easier and quicker to tell if an argument supports a conclusion compared to deciding whether an argument supports another argument. 

\paragraph{ARC systems} A linear SVM classifier that is trained on a diverse set of features provides competitive performance on \textsc{Essay} \cite{stab2017parsing}. A subsequent joint global graph optimization step, similarly to \cite{peldszus2015joint,hou2017argument}, yields no further improvement for classifying the relations in this data. The SVM classifier incorporates features extracted from the EAU spans as well as their context (e.g., leading or trailing words). On \textsc{Essay-Content}, where systems only see the EAU clause spans, the performance of the SVM suffers a loss of more than 10 pp.\ macro F1 \cite{opitzdissec} -- an analysis indicates that the SVM focuses immoderately on features extracted from the EAU context and tends to neglect their actual content. This underpins the need for argumentative relation classification systems with deeper understanding of argumentation, i.e., systems that base their prediction on the actual content of two EAUs -- the method we present in this paper aims at this. 

The first neural approach for ARC \cite{cocarascu2017identifying} proposes a neural network with a Siamese structure \cite{koch2015siamese,mueller2016siamese,cocarascu2017identifying}. By means of a shared weight space it projects source and target EAU to a joint distributional vector space. Finally, it classifies the vector offset using a softmax-function. The authors conduct experiments on a data set which comprises texts about movies, technology and  politics.

A similar model has been adopted recently where (symbolic) knowledge from large background knowledge-graphs is injected into the Siamese model by concatenating highly abstracted  multi-hop knowledge paths to the source-target offset \cite{kobbe_et_al:OASIcs:2019:10372}. 
Although there are consistent gains observed by including the knowledge, the gains appear to be relatively small. In this aspect, we believe that incorporating knowledge of the right form could make it possible to further enhance the system we propose in this paper. However, as of now, it is an active topic of discussion \textit{whether} (symbolic) background knowledge may help in automatic argumentation and, even more so, \textit{which} (form of) knowledge would be needed.

\paragraph{Computational argument mining and analysis} Argumentation is ubiquitous and argumentative structures can be recovered from a broad spectrum of texts. For example, they can be recovered from online dialogue \cite{swanson2015argument,budzynska2014towards} and scientific research articles \cite{lauscher2018arguminsci,lauscher-etal-2018-argument}, where, e.g., researchers may directly or indirectly convey arguments for why some method is better than another. By now, there exists a substantial body of research publications covering a variety of argument analysis topics. For a general overview, we refer the reader to \newcite{lippi2016argumentation}  and \newcite{peldszus2013argument}.

\paragraph{Plausibility ranking} Another task that can be addressed as a text plausibility ranking task is the resolution of difficult pronouns in the \textit{Winograd Schema Challenge} \cite{Levesque:2012:WSC:3031843.3031909,opitz-frank-2018-addressing}. To resolve shell nouns and abstract anaphora (e.g., `I like \textit{that}'.) \newcite{marasovic-etal-2017-mention} utilize syntactic patterns to gather plausible candidate resolutions from a background corpus in order to extend the scarce training data. 

\section{Context reconstruction and model}

In this section, we first propose a simple reconstruction trick which allows us to build minimal pairs of plausible and implausible argumentative texts. Then, we describe a Siamese neural sequence ranking model which addresses the task of ranking texts according to their plausibility.
\label{sec:rec}
\paragraph{Constructing plausible and implausible argumentative discourse contexts} Consider two EAU clauses $a_2$ (source) and $a_1$ (target) where we need to decide whether $a_2$ \textit{supports} $a_1$ or $a_2$ \textit{attacks} $a_1$. In the absence of contextual discourse clues\footnote{To name just one situation: consider a cross-document relation classification setup where $a_1$ stems from a different document than $a_2$. Any specific textual discourse context would not only be more or less unimportant, but also bears the potential to confuse the system.}, a system must learn to predict this relation by considering the semantic content of $a_1$ and $a_2$. We approach this task by offering two alternative context reconstructions and asking our model in what context $a_1$ and $a_2$ are more likely to appear. More precisely, our reconstruction trick is as follows: 

\begin{description}
    \item[(a)] $a_1$. Additionally, $a_2$ .
    \item[(b)] $a_1$. Admittedly, $a_2$.
    \label{ex:1}
\end{description}

where \textbf{(a)} signals that two argumentative units likely stand in a \textit{support}-relation and \textbf{(b)} signals the opposite (`attack'). In our experiments (Section \ref{sec:ex}), we also examine other possible discourse connectors for our reconstruction (e.g., moreover/however). From here, we ask our model which of the two reconstructions leads to a more plausible `reading': \textbf{(a)} or \textbf{(b)}? E.g., consider the \textit{cannabis}-example from Section \ref{sec:intro}; applying our reconstruction trick yields the following \textit{implausible-plausible} minimal pair $(r^-,r^+)$:

\begin{description}
    \item[(3a)] [\textbf{$_{r^-}$} \textit{Use of marijuana causes chronic bronchitis and airflow obstruction. \textbf{Additionally}, cannabis does not need to be smoked to receive its potential health benefits.]}
    \item[(3b)] [\textbf{$_{r^+}$} \textit{Use of marijuana causes chronic bronchitis and airflow obstruction. \textbf{Admittedly}, cannabis does not need to be smoked to receive its potential health benefits.]}
    \label{ex:2}
\end{description}

Clearly, \textbf{(3b)} constitutes a more plausible reconstruction compared with \textbf{(3a)}. Exactly this is what we desire our model to learn: assessing the fine-grained differences between two texts which differ in only one phrase. This phrase, however, determines whether the text in its entirety is \textit{implausible} or \textit{plausible}.

\subsection{Loss and model}

\paragraph{Ranking loss} We argue that a ranking approach (\textit{which reading is more plausible?}) is more suitable for addressing our problem compared with a classification approach (\textit{plausible vs.\ implausible}). The reason is that ranking allows for a more relaxed and graded notion of textual plausibility: we want the model to \textit{prefer} one variant and not to \textit{choose} one variant. This is accomplished by reducing the margin ranking loss on the training data $\{(r^+_i,r^-_i)\}_{i=1}^n$:

\begin{equation}
\label{eq:loss}
    \mathcal{L_\theta} = \frac{1}{n}\sum_{i=1}^{n}\bigg[ 1 - score_\theta(r^+_i) + score_\theta(r^-_i)\bigg],
\end{equation}

where $score(\cdot)$ is a plausibility prediction model parameterized by $\theta$. The plausibility-prediction model which we use is described in detail in the following paragraphs. Since $\mathcal{L_\theta}$ is differentiable with respect to the model's parameters $\theta$, we can learn them with gradient descent.

\begin{figure}
    \centering
    \includegraphics[trim=650 500 1 10,clip,width=1.2\linewidth]{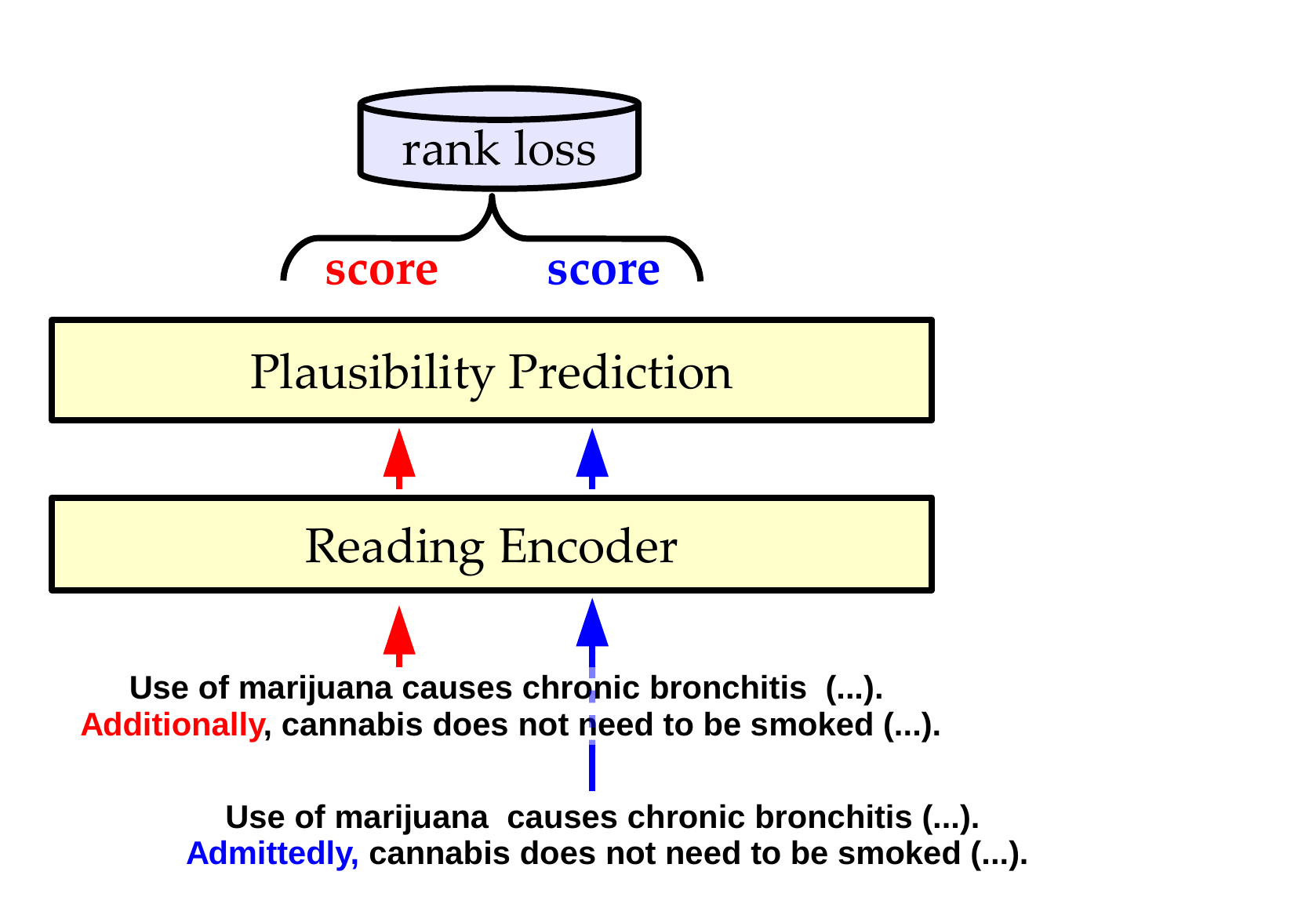}
    \caption{Siamese model outline. Two competing reading-reconstructions are fed through a Siamese encoder (Reading Encoder). The vectors are then mapped by means of a (Siamese) linear combination and a selu-activation onto two corresponding plausibility scores (Plausibility Prediction). By reducing the ranking loss, we force the model to assign higher scores to more plausible readings.}
    \label{fig:mod}
\end{figure}
\paragraph{Model overview} We desire the $score(\cdot)$ function to return a number $p \in \mathbb{R}$ reflecting the plausibility of a text sequence made up of words $w_1,...w_n$. In our case, this function is instantiated with (i) a Siamese reading encoder (\textit{Reading Encoder}, Figure \ref{fig:mod}) and a Siamese plausibility prediction layer for producing a plausibility score for any given text (\textit{Plausibility Prediction}, Figure \ref{fig:mod}). Now, we will describe these two components more closely. 

\paragraph{Reading encoder} First, we use a contextual language model\footnote{We use BERT \cite{devlin-etal-2019-bert} to infer the contextual embeddings. In our ablation experiments, we also present results based on ELMo embeddings \cite{peters-etal-2018-deep}} to infer a sequence of word embeddings: $e_1,...,e_n$, which correspond to words $w_1,...w_n$. Here, we hope that already the contextual language model provides statistical information indicating whether a specific word sequence may be considered as rather plausible or rather implausible (`inductive bias'). The sequence of word embeddings $e_1,...,e_n$ is further multiplied by a sequence of positive indicator coefficient embeddings: $e_1\cdot c_1,...,e_n\cdot c_n$.\footnote{Similar to \newcite{opitz-frank-2019-argument}.} This allows the model to learn to better distinguish between the source, target and the connector text (we learn three corresponding indicator embeddings). The resulting sequence is further processed by (ii) a Bi-LSTM \cite{hochreiter1997long} to construct hidden states $H=h_1,...,h_n$ (we concatenate hidden states of forward and backward read) and (iii) a four-headed scaled dot-product self-attention mechanism \cite{NIPS2017_7181}, where in our case we use $H=Q=K=V$:

\begin{align*}
    Heads (Q,K,V) &= [head_1;...,head_4]W^O\\
    head_i &= Attention(QW_i^Q,KW_i^K,VW_i^V)\\
    Attention(Q, K, V) &= softmax(\frac{QK^T}{\sqrt{d_k}})V,
 \end{align*}


where $W_{(\cdot)}^{(\cdot)}$ are parameters of the model. Finally, we compute a weighted average of the final sequence of hidden states to construct a vectorized reading representation $v$ \cite{felbo-etal-2017-using}:

\begin{align*}
e_t &=   Heads(\cdot)_tW^A\text{~~~~~~~~~~} a_t = \frac{exp(e_t)}{\sum_{i=1}^{T}{exp(e_i)}}\\
v &= \sum_{i=1}^{T}{a_i Heads(\cdot)_i},
\end{align*}

where $Heads(\cdot)_t$ is the vector corresponding to time step $t$ computed by the previous scaled dot-product attention step and $v$ is a final vectorized representation of the input reading.

\paragraph{Plausibility prediction} At plausibility prediction time, the vector representation $v$, which we obtained by the previous step, is mapped to a single score by means of a linear combination with a weight vector. Lastly, a selu-function \cite{DBLP:journals/corr/KlambauerUMH17} produces the desired plausibility-score:
\begin{equation}
 p=selu(v^Tw).
\end{equation}

This score, computed once for each of the two competing reconstructions, allows a comparison with respect to their (predicted) plausibility. For our ARC experiments, where we desire a final classification, we predict the argumentative relation class by inspecting the discourse connector of the reconstruction which obtains a higher plausibility score. E.g., if $score(EAU_1, additionally, EAU_2) \geq score(EAU_1, admittedly, EAU_2)$ we predict the argumentative `support' relation -- otherwise we predict the `attack' relation.

\section{Experiments}

We begin this section by describing the experimental setup used to evaluate our neural plausibility ranker. Next, we present our main results and finally perform several analyses and study the effects of ablating model components.
\label{sec:ex}
\subsection{Setup}

\paragraph{Discourse links} To construct plausible and implausible texts, we experiment with eight different discourse connectors which have the potential to `signal' argumentative relation types. They make up, in total, four minimal pairs (Table \ref{tab:plim}). 
\begin{table}
    \centering
    \begin{tabular}{l|ll}
    \toprule
        abbreviation\ & `support' & `attack' \\
    \midrule
        A/A & Additionally, & Admittedly,\\
       A/D & I agree, & I disagree, \\
       M/H& Moreover, &  However, \\
       Y/N & Yes, &  No, \\
       \bottomrule
    \end{tabular}
    \caption{Argumentative discourse connector sentence adverbials and the argumentative relation class which they are likely to signal.}
    \label{tab:plim}
\end{table}


\paragraph{Data}\label{par:data} We use the student essay corpus v02 \cite{stab2017parsing} in two versions: \textsc{Essay} and \textsc{Essay-Content}. What is common to both is that they  contain data from the same 402 argumentative essays written by students about a variety of topics. The essays have been annotated with, i.a.\, spans of argumentative units and their relations with each other (support vs.\ attack). Since only the argumentative clauses have been annotated, we can clean EAUs from their discourse context, which yields \textsc{Essay-Content}. For example, consider \textit{$EAU_1$. To add on this, $EAU_2$}. While in \textsc{Essay}, a system is allowed to see EAU-surrounding tokens (\textit{to add on this}), in \textsc{Essay-Content}, systems are allowed to see only the spans of the EAUs to predict their relation (i.e., $EAU_{1}, EAU_{2}$).  In the easy case, \textit{to add on this} may be enough to predict a support relation with high confidence and accuracy without even seeing the content of the EAUs -- in the hard case, however, a system must learn to assess the actual content of the premises. In \textsc{Essay-Content}, the performance of the feature-based SVM described by \newcite{stab2017parsing} drops by more than 23\% macro F1 compared to the standard setup (\textsc{Essay}) where shallow discourse context is accessible \cite{opitzdissec}.

\paragraph{Baselines} We display the results of a competitive feature based SVM. It requires, i.a., syntactic parsing, constituency-tree sentiment annotation \cite{socher-etal-2013-recursive} and discourse parsing \cite{pdtbdiscou} as pre-processing steps \cite{stab2017parsing,opitzdissec}. In contrast, our method does not depend on any pre-processing.

\paragraph{Model instantiation} For each possible minimal pair, we instantiate a different model based on the pre-trained BERT model (the BERT model remains fixed during optimization). More specifically, we infer the word embeddings and average over the last four layers to produce a sequence of vectors with 1024 dimensions. Forward and backward LSTM have 256 neurons each. For development purposes we split off 1149 examples from the training data. The rank loss (Eq. \ref{eq:loss}) is minimized by performing stochastic gradient descent with Adam \cite{kingma2014adam}\footnote{The learning rate is set to 0.001, the mini-batch size to 64 and the maximum number of epochs to 25.}. After each epoch, the model is evaluated on the development data. Finally, we select the parameters from the epoch with maximum F1 score on the development data.

In our tables, each model is denoted $ArgRanker_{dcs}$ where $dcs$ indicates which pair of discourse connectors was used for reconstruction. $ArgRanker_{vote}$ denotes a model where we aggregate the predictions over the four different minimal-pair single models (`ensemble model'). All results are averaged over five runs. 

\subsection{Results}

\begin{table}
    \centering
    \scalebox{0.9}{
    \begin{tabular}{l|l|l}
    \toprule
    System&\textsc{Essay}&\textsc{Essay-Content}\\
         
         \midrule
         majority baseline & 47.8 & 47.8 \\
         SVM with features & 68.0 & 57.3 \\
         \midrule
         ArgRanker$_{A/A}$& 67.2$^{\pm 1.0}$ & 58.6$^{\pm 1.4}$ \\
         ArgRanker$_{A/D}$& 69.2$^{\pm 2.4}$ &\textbf{59.2$^{\pm 0.7}$ } \\
         ArgRanker$_{M/H}$& 68.8$^{\pm 1.7}$ &\underline{\textbf{63.8$^{\pm 2.1}$}}  \\
         ArgRanker$_{Y/N}$& 67.3$^{\pm 0.8}$ & 58.3$^{\pm 1.8}$  \\
         \midrule
         ArgRanker$_{vote}$&\underline{\textbf{70.9$^{\pm 0.7}$}} &\textbf{60.7$^{\pm 1.7}$}\\
         \bottomrule
    \end{tabular}}
    \caption{Macro F1 results. \underline{underlined}: best result; \textbf{bold}: improves against SVM withstanding standard deviation.}
    \label{tab:mainres}
\end{table}
 
 \paragraph{Macro F1 results}  Table \ref{tab:mainres} lists the macro F1 results\footnote{Macro F1 in our case is defined as the unweighted mean over the F1 scores for our two classes.} of our experiments. 
 
 On \textsc{Essay}, our method is competitive with the SVM that relies on extensive pre-processing. On \textsc{Essay-Content}, where models are forced to learn to assess the content of EAUs, our method outperforms the feature-based SVM across all configurations. The best performance on this data is provided by ArgRanker$_{M/H}$, which is trained on \textit{Moreover-However} reconstructions (+6.5 pp.\ macro F1, relative improvement: 11\%). Our ensemble model ArgRanker$_{vote}$, which aggregates the predictions of the individual ArgRankers in a simple vote, achieves an improvement of +3.4 pp.\ macro F1 (relative improvement: 6\%).
 
\paragraph{More detailed results} Table \ref{tab:sublassres} indicates that our method offers other advantages besides raw macro F1 gains. The very rare \textit{attack}-class is detected with a much greater precision compared with the SVM. The difference can range from an improvement of 5.6 pp.\  (ArgRanker$_{Y/N}$, relative improvement: 28\%) up to a maximum improvement of 31.4 pp. (ArgRanker$_{vote}$, relative improvement: 157\%). With such a large increase in precision, one might expect a drop in recall -- however, this is only the case to a very small extent. 
\begin{table}
    \centering
    \scalebox{0.72}{
    \begin{tabular}{lllll}
    \toprule
         &\multicolumn{2}{c}{\textit{Attack}}&\multicolumn{2}{c}{\textit{Support}}\\
         System&Precision & Recall &Precision &Recall\\
         \midrule
         majority & 0.00 & 0.00 &8.0 & \underline{100.0} \\
         SVM with features &20.0 & 22.9 & 93.0 &91.8 \\
         \midrule
         ArgRanker$_{A/A}$& \textbf{31.0$^{\pm 5.9}$} & 18.2$^{\pm 2.1}$ & 92.9$^{\pm 0.1}$ & \textbf{96.2$^{\pm 1.1}$ }\\
         ArgRanker$_{A/D}$& \textbf{28.0$^{\pm 4.4}$} &22.8$^{\pm 3.2}$  & 93.1$^{\pm 0.2}$& \textbf{94.5$^{\pm 1.6}$} \\
         ArgRanker$_{M/H}$& \textbf{39.9$^{\pm 9.4}$} &\underline{\textbf{30.0$^{\pm 6.8}$}}& \textbf{\underline{93.8$^{\pm 0.6}$}} & \textbf{95.5$^{\pm 2.4}$ } \\
         ArgRanker$_{Y/N}$& 25.6$^{\pm 5.8}$ & 23.7$^{\pm 5.1}$& \textbf{93.1$^{\pm 0.3}$}& 93.0$^{\pm 3.3}$ \\
         \midrule
         ArgRanker$_{vote}$&\underline{\textbf{51.4$^{\pm 7.3}$}} &17.7$^{\pm 3.4}$ & 93.0$^{\pm 0.2}$& \textbf{98.4$^{\pm 0.7}$} \\
         \bottomrule
    \end{tabular}}
    \caption{Precision and recall scores for each class on \textsc{Essay-Content}. \underline{underlined}: best result; \textbf{bold}: improves against SVM withstanding standard deviation.}
    \label{tab:sublassres}
\end{table}
 The greatest drop in recall is incurred by ArgRanker$_{vote}$ (-5.2 pp.) and thus can be said to lie in the shadow of its precision gains (+31.4 pp.). Moreover, when we use the discourse connector minimal pairs A/D and M/H, our model outperforms the SVM in the \textit{attack}-class both in precision and recall. Most notably, when we instantiate our reconstructions with \textit{Moreover/However}, we see a large gain in precision (+19.9 pp., relative improvement: 99.5\%) but also an observable gain in recall (+7.1 pp., relative improvement: 31.0\%). 
 
 With regard to the majority class (\textit{support}), we make two observations: (i) precision-wise, all of our models outperform or are on par with the SVM; (ii) recall-wise, all of our models outperform the SVM. The greatest gain in recall for \textit{support} is achieved by ArgRanker$_{vote}$ (+6.6 pp.).

\subsection{Ablation experiments and analysis}
\label{sec:abl}
\begin{table*}
    \centering
    \begin{tabular}{l|l|lll}
    &\multicolumn{4}{c}{model configuration}\\
    \toprule
    System&basic&ELMo&-coeff.\ & -att.\   \\
         \midrule
         majority  & 47.8& -& -& - \\
         SVM \cite{stab2017parsing,opitzdissec} & 57.3 & -& -& - \\
         \midrule
         ArgRanker$_{A/A}$& \textbf{58.6$^{\pm 1.4}$ }& 55.7$^{\pm 1.6}$&57.4$^{\pm 2.1}$&58.3$^{\pm 1.2}$\\
         ArgRanker$_{A/D}$ &59.2$^{\pm 0.7}$ & \textbf{60.2$^{\pm 2.2}$} &59.6$^{\pm 2.1}$&56.2$^{\pm 1.9}$\\
         ArgRanker$_{M/H}$ &\textbf{63.8$^{\pm 2.1}$} &59.4$^{\pm 2.5}$&61.1$^{\pm 1.0}$ &60.7$^{\pm 1.9}$ \\
         ArgRanker$_{Y/N}$& \textbf{58.3$^{\pm 1.8}$} & 57.6$^{\pm 1.7}$ &57.7$^{\pm 1.2}$&58.2$^{\pm 3.0}$\\
         \midrule
        
         ArgRanker$_{vote}$ &\textbf{60.7$^{\pm 1.7}$}&59.5$^{\pm 1.9}$&60.2$^{\pm 2.2}$ & 56.2\\
         \midrule
         ArgRanker$_{-discourse}$&  57.3$^{\pm 0.4}$ & \textbf{63.6$^{\pm 1.3}$} & 57.3$^{\pm 2.8}$ & 54.5\\
    
         \bottomrule
    \end{tabular}
    \caption{Ablation experiments: Macro F1 results on \textsc{Essay-Content}. \textit{ArgRanker$_{-discourse}$}: a system where we replace the natural discourse connectors with `linguistically meaningless' placeholders (i.e., support: `+', attack: `-' instead of, e.g., support: `Moreover', attack: `However'). \textit{ELMo}: we use ELMo instead of BERT; \textit{-coeff.}: we abstain from learning source-target specific coefficients; \textit{-att.}: we ablate the self-attention and use the last states of the Bi-LSTM (concatenation of each read) for prediction.}
    \label{tab:abl}
\end{table*}

\begin{figure}
    \centering
    \includegraphics[scale=0.54]{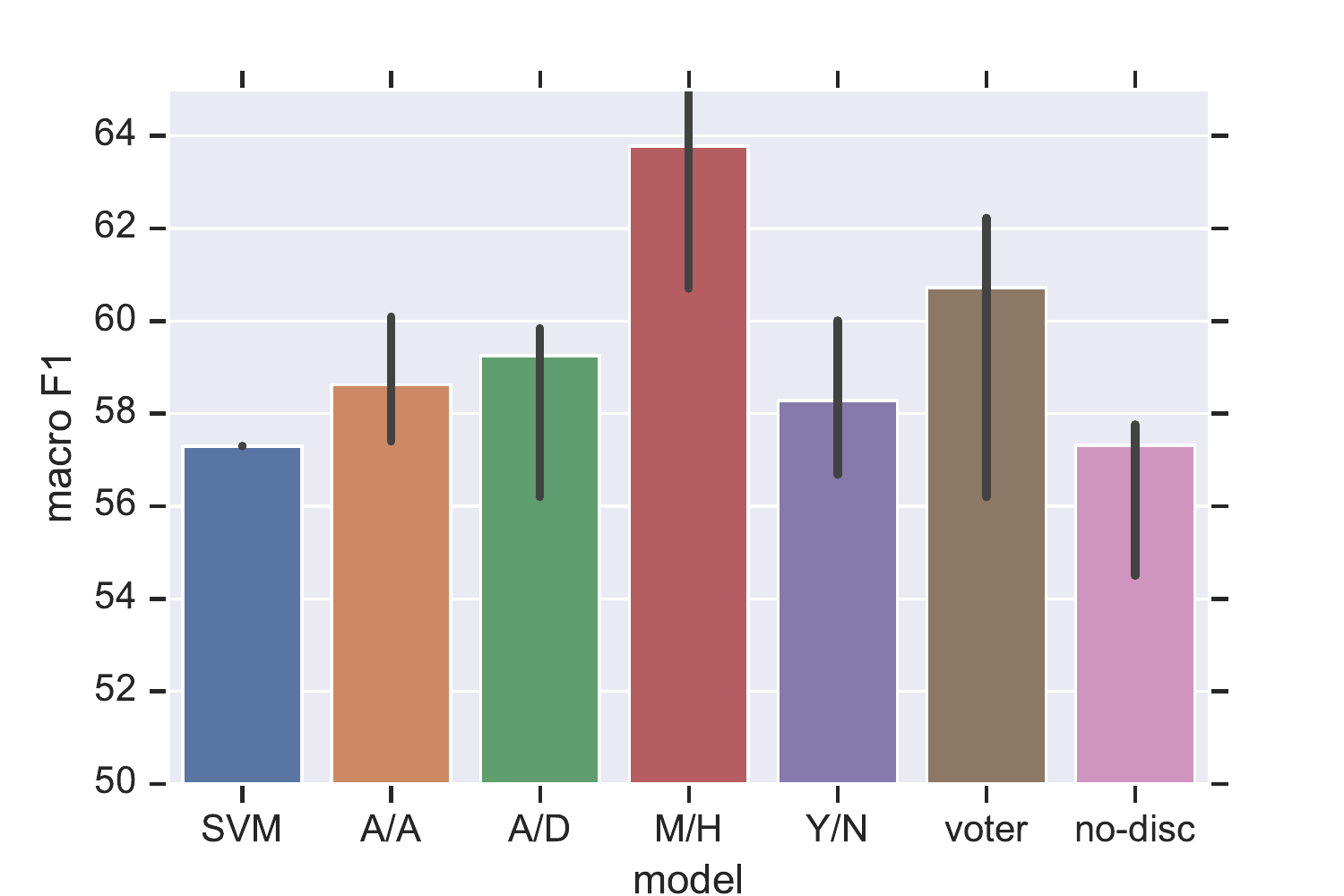}
    \caption{Scores for different models using BERT embeddings and SVM (left column) on \textsc{Essay-Content}. Reconstruction with \textit{Moreover/however} offers the largest improvement. Non-linguistically motivated connectors  result in reduced performance (\textit{`+'/`-'}: \textit{no-disc}, right column). }
    \label{fig:as}
\end{figure}

\paragraph{Linguistically motivated discourse reconstruction} What is the outcome of  instantiating the discourse reconstructions with `meaningless' connectors? I.e., instead of instantiating the attack/support context with linguistically motivated connectors, such as, e.g., \textit{I agree/I disagree}, we instantiate the contexts with the meaningless tokens `+' and `-'. On one hand, this means that the new discourse configuration is still discriminative (either supporting or attacking). On the other hand, however, the discriminating reconstruction is not any more linguistically motivated. Thus, we hypothesize that the 
linguistically motivated reconstructions better `trigger' the contextual BERT model into giving a useful inductive bias about whether a certain reading is plausible or not. 

From Table \ref{tab:abl} and Figure \ref{fig:as}, we see that, indeed, our model functions better when provided with linguistically motivated reconstructions instead of the non-linguistically motivated reconstruction (Figure \ref{fig:as}: columns \textit{A/A, A/D, M/H, Y/N} vs.\ bottom row in Table \ref{tab:abl} and \textbf{\textcolor{blush}{right column}} in Figure \ref{fig:as}). This holds true across all model configurations and all linguistically motivated discourse connector pairs.\footnote{An exception constitutes the model based on ELMo embeddings, which appears to work better when provided with the non-linguistically motivated connector pair.} 

More specifically, we find that the \textit{Moreover/However} reconstruction appears to offer the most useful inductive bias (\textbf{\textcolor{cornellred}{middle column}}, Figure \ref{fig:as}). Our \textit{ArgRanker} based on this reconstruction outperforms all other configurations by more than 4 pp.\ macro F1 (compared with \textit{Agree/Disagree}) and more than 6 pp.\ macro F1 compared with the non-linguistically motivated reconstruction. One reason could be located in the fact that BERT was trained, i.a., on the Wikipedia corpus: we compute a simple word frequency statistic over this corpus and see that the terms \textit{Moreover} and \textit{However} appear more frequently in this corpus (e.g., \textit{however}: appr.\ 29,900,000 occurrences) than, e.g., \textit{Admittedly} (appr.\ 17,000 occurrences). Also, by manually inspecting a small amount of occurrences in Wikipedia, we find that \textit{moreover} and \textit{however} tend to occur in more `argumentative' contexts, or, at least, connect two discourse units in a contrasting (\textit{however}) or supporting (\textit{moreover}) way. On the other hand, e.g., \textit{I agree} tends to occur in less argumentative contexts, such as in \textit{I agree to the terms of service}. We believe that contextual language models trained on interactive discourse texts (e.g., online discussion platforms) instead of encyclopedic texts would greatly help to provide our model with better embeddings in the situations where we want to compose plausible and implausible texts by means of more `interactive' connectors (\textit{I agree/I disagree}; \textit{Yes/No}; etc.).

\begin{figure*}
\begin{subfigure}{.33\textwidth}
  \centering
   \frame{\includegraphics[width=\linewidth]{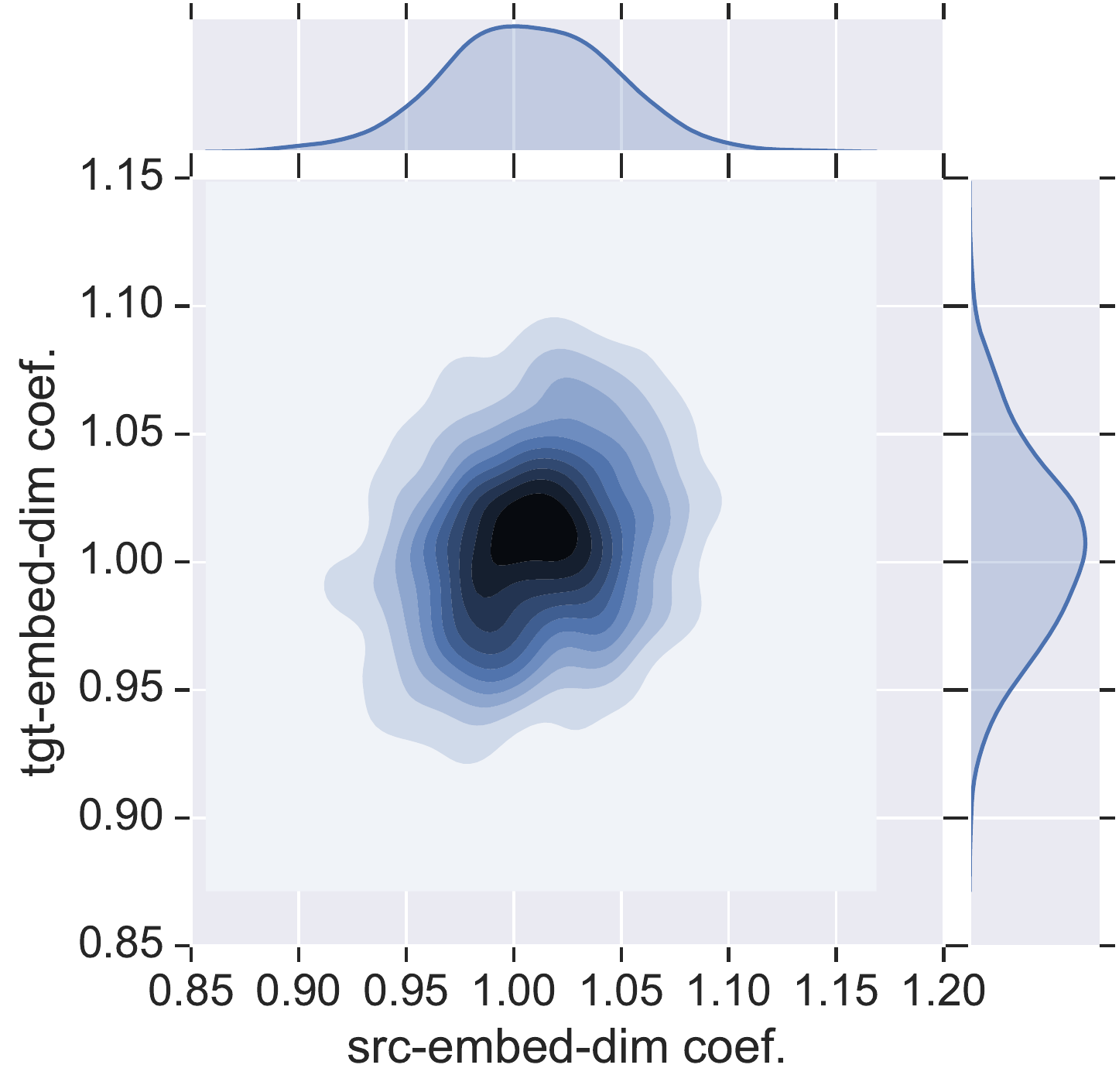}}
  \caption{Target vs.\ source coefficients}
  \label{fig:sfig1}
\end{subfigure}%
\begin{subfigure}{.33\textwidth}
  \centering
  \frame{\includegraphics[width=\linewidth]{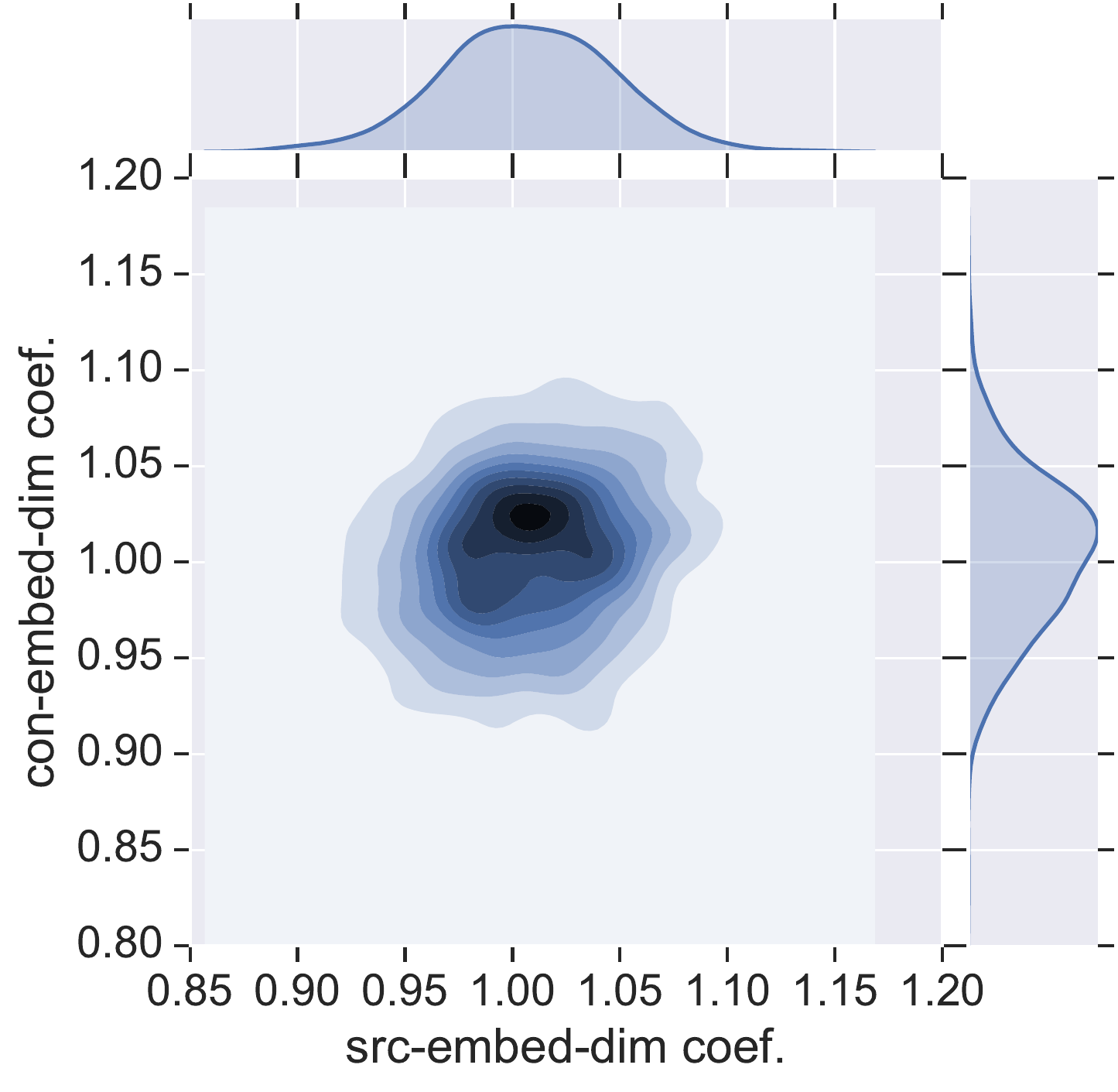}}
  \caption{Connector vs.\ source coefficients}
  \label{fig:sfig2}
\end{subfigure}%
\begin{subfigure}{.33\textwidth}
  \centering
   \frame{\includegraphics[width=\linewidth]{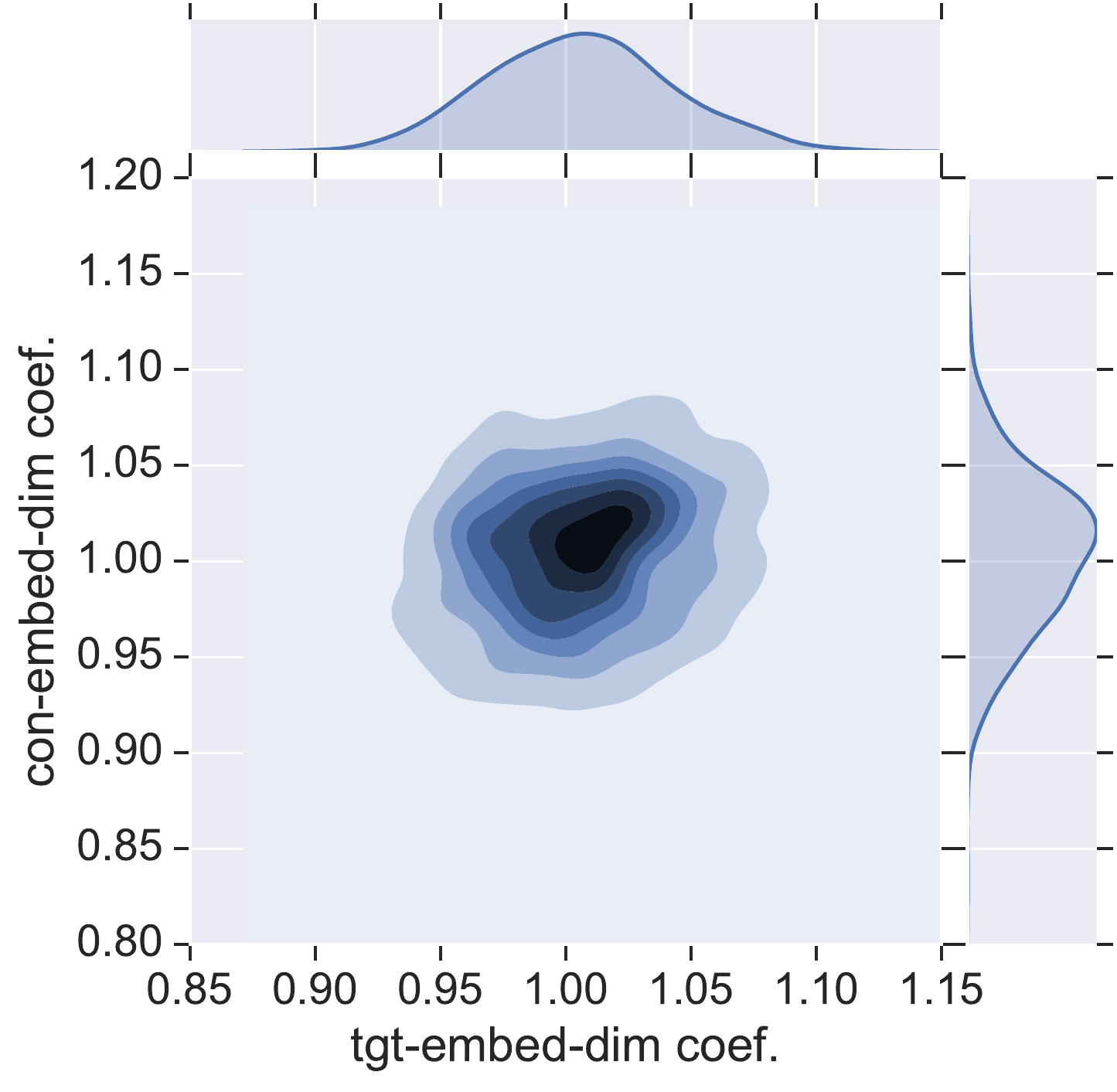}}
  \caption{Connector vs.\ target coefficients}
  \label{fig:sfig3}
\end{subfigure}
  \caption{Investigation of the three contextual coefficient-embeddings which our model has learned. The coefficients are initialized with ones and assume, during training, a Gaussian-like distribution. By all appearances, the model uses some coefficients to `deflate' the impact of an embedding dimension in, e.g., the text corresponding to the source EAU, while `inflating' the impact of an embedding dimension in, e.g., the  text corresponding to the target EAU (Figure \ref{fig:sfig1}, regions on the upper left).}
\label{fig:figcoef}
\end{figure*}

\paragraph{BERT vs.\ ELMo} In our first experiment, we replace the BERT embeddings with ELMo embeddings -- we want to `probe' which of the two embedding generators is better suited to rank argumentative texts according to their plausibility. First, we see that ELMo embeddings provide better performance than the feature based baseline, with one exception: ArgRanker$_{A/A}$, where we reconstruct contexts by inserting \textit{Additionally} and \textit{Admittedly} (Table \ref{tab:abl}, \textbf{ELMo}). Second, the ELMo embeddings in most cases fall short in comparison to BERT embeddings -- again, however, with one exception: ArgRanker$_{-discourse}$, which does not use the linguistically motivated reconstruction. 

\paragraph{Indicator embedding coefficients} Now, we want to investigate if learning coefficients to better distinguish between source and target has helped our model. Recall, that the three coefficient indicator embeddings correspond to \textit{source/target EAU span} and the \textit{discourse connector span} and allow the model to highlight certain word embedding indices differently with respect to these three spans.  For most connector pairs, learning the coefficients helps and their ablation leads to a performance drop (Table \ref{tab:abl}, \textbf{-coeff}; e.g.\ ArgRanker$_{M/H}$: -2.7 pp.\ macro F1). 

Finally, we plot the learnt coefficients of the three different indicator embeddings against each other to analyze their appearance after training. Figure \ref{fig:figcoef} displays all values from all discourse connector parameterizations $\cdot/\cdot$ of ArgRanker$_{\cdot/\cdot}$. More specifically, we are interested in the following question: \textit{Have we learned that certain contextual word embedding indices are important to inflate (deflate) with respect to the source or the target?} From inspecting Figure \ref{fig:figcoef}, we see that this appears to be the case. For example, there is a set of embedding indices where coefficients are used to magnify the corresponding values in the target EAU and deflate them in the source EAU (Figure \ref{fig:sfig1}, top left region) -- while for another set of embedding indices the opposite is true  (Figure \ref{fig:sfig1}, bottom right). Furthermore, the learnt coefficients have assumed a normal-like distribution after training (distribution plots on the sides of Figures \ref{fig:sfig1}, \ref{fig:sfig2}, \ref{fig:sfig3}).

\paragraph{Self-attention} Finally, we want to investigate the effect of ablating the self-attention mechanisms from our model. More precisely, we predict the plausibility scores based on a concatenation of the last state of forward and backward read of the Bi-LSTM. Throughout all different discourse reconstruction strategies, we see drops in performance (Table \ref{tab:abl}, \textbf{-att}). However, while we see observable drops in some cases (ArgRanker$_{vote}$: -4.5 pp.\ macro F1), they are comparatively small in other cases (ArgRanker$_{Y/N}$: -0.1 pp.).

\section{Conclusion}

We have treated argumentative relation classification in a new light, as a task where we learn to rank candidate texts according to their plausibility. To this aim, we have proposed a simple reconstruction trick which allows us to embed source and target argumentative units into plausible and implausible argumentative discourse contexts. In order to learn to rank such texts according to their plausibility, we have adapted a neural Siamese ranking model. Our experiments on an established data set have shown that the approach is competitive with previous work albeit it does not require pre-processing. 
In the `content-based' setting -- which is more difficult because models cannot base their decisions on shallow clues in the discourse context -- the method outperforms previous work by a considerable margin. In particular with respect to the scarce class \textit{attack} we observed substantial improvements in precision. 


\section*{Acknowledgments}

We are grateful to Anette Frank for valuable discussions and feedback on an earlier draft of this paper. This work has been supported by the  Leibniz  ScienceCampus ``Empirical  Linguistics  and  Computational  Language  Modeling'', supported by the Leibniz Association grant no.\  SAS-2015-IDS-LWC  and by the  Ministry of Science, Research, and Art of Baden-W\"urttemberg.

\bibliographystyle{konvens2019}
\bibliography{bibl.bib}

\begin{thebibliography}{}

\bibitem[\protect\citename{Besnard and Hunter}2014]{besnard2014constructing}
Philippe Besnard and Anthony Hunter.
\newblock 2014.
\newblock Constructing argument graphs with deductive arguments: a tutorial.
\newblock {\em Argument \& Computation}, 5(1):5--30.

\bibitem[\protect\citename{Budzynska \bgroup et al.\egroup
  }2014]{budzynska2014towards}
Katarzyna Budzynska, Mathilde Janier, Juyeon Kang, Chris Reed, Patrick
  Saint-Dizier, Manfred Stede, and Olena Yaskorska.
\newblock 2014.
\newblock Towards argument mining from dialogue.
\newblock In {\em COMMA}, pages 185--196.

\bibitem[\protect\citename{Cocarascu and Toni}2017]{cocarascu2017identifying}
Oana Cocarascu and Francesca Toni.
\newblock 2017.
\newblock Identifying attack and support argumentative relations using deep
  learning.
\newblock In {\em Proceedings of the 2017 Conference on Empirical Methods in
  Natural Language Processing}, pages 1374--1379.

\bibitem[\protect\citename{Devlin \bgroup et al.\egroup
  }2019]{devlin-etal-2019-bert}
Jacob Devlin, Ming-Wei Chang, Kenton Lee, and Kristina Toutanova.
\newblock 2019.
\newblock {BERT}: Pre-training of deep bidirectional transformers for language
  understanding.
\newblock In {\em Proceedings of the 2019 Conference of the North {A}merican
  Chapter of the Association for Computational Linguistics: Human Language
  Technologies, Volume 1 (Long and Short Papers)}, pages 4171--4186,
  Minneapolis, Minnesota, June. Association for Computational Linguistics.

\bibitem[\protect\citename{Felbo \bgroup et al.\egroup
  }2017]{felbo-etal-2017-using}
Bjarke Felbo, Alan Mislove, Anders S{\o}gaard, Iyad Rahwan, and Sune Lehmann.
\newblock 2017.
\newblock Using millions of emoji occurrences to learn any-domain
  representations for detecting sentiment, emotion and sarcasm.
\newblock In {\em Proceedings of the 2017 Conference on Empirical Methods in
  Natural Language Processing}, pages 1615--1625, Copenhagen, Denmark,
  September. Association for Computational Linguistics.

\bibitem[\protect\citename{Hochreiter and Schmidhuber}1997]{hochreiter1997long}
Sepp Hochreiter and J{\"u}rgen Schmidhuber.
\newblock 1997.
\newblock Long short-term memory.
\newblock {\em Neural computation}, 9(8):1735--1780.

\bibitem[\protect\citename{Hou and Jochim}2017]{hou2017argument}
Yufang Hou and Charles Jochim.
\newblock 2017.
\newblock Argument relation classification using a joint inference model.
\newblock In {\em Proceedings of the 4th Workshop on Argument Mining}, pages
  60--66.

\bibitem[\protect\citename{Kingma and Ba}2014]{kingma2014adam}
Diederik~P Kingma and Jimmy Ba.
\newblock 2014.
\newblock Adam: A method for stochastic optimization.
\newblock {\em arXiv preprint arXiv:1412.6980}.

\bibitem[\protect\citename{Klambauer \bgroup et al.\egroup
  }2017]{DBLP:journals/corr/KlambauerUMH17}
G{\"{u}}nter Klambauer, Thomas Unterthiner, Andreas Mayr, and Sepp Hochreiter.
\newblock 2017.
\newblock Self-normalizing neural networks.
\newblock {\em CoRR}, abs/1706.02515.

\bibitem[\protect\citename{Kobbe \bgroup et al.\egroup
  }2019]{kobbe_et_al:OASIcs:2019:10372}
Jonathan Kobbe, Juri Opitz, Maria Becker, Ioana Hulpus, Heiner Stuckenschmidt,
  and Anette Frank.
\newblock 2019.
\newblock {Exploiting Background Knowledge for Argumentative Relation
  Classification}.
\newblock In Maria Eskevich, Gerard de~Melo, Christian F{\"a}th, John~P.
  McCrae, Paul Buitelaar, Christian Chiarcos, Bettina Klimek, and Milan
  Dojchinovski, editors, {\em 2nd Conference on Language, Data and Knowledge
  (LDK 2019)}, volume~70 of {\em OpenAccess Series in Informatics (OASIcs)},
  pages 8:1--8:14, Dagstuhl, Germany. Schloss Dagstuhl--Leibniz-Zentrum fuer
  Informatik.

\bibitem[\protect\citename{Koch \bgroup et al.\egroup }2015]{koch2015siamese}
Gregory Koch, Richard Zemel, and Ruslan Salakhutdinov.
\newblock 2015.
\newblock Siamese neural networks for one-shot image recognition.
\newblock In {\em ICML deep learning workshop}, volume~2.

\bibitem[\protect\citename{Lauscher \bgroup et al.\egroup
  }2018a]{lauscher2018arguminsci}
Anne Lauscher, Goran Glava{\v{s}}, and Kai Eckert.
\newblock 2018a.
\newblock Arguminsci: A tool for analyzing argumentation and rhetorical aspects
  in scientific writing.
\newblock In {\em Proceedings of the 5th Workshop on Argument Mining}, pages
  22--28.

\bibitem[\protect\citename{Lauscher \bgroup et al.\egroup
  }2018b]{lauscher-etal-2018-argument}
Anne Lauscher, Goran Glava{\v{s}}, and Simone~Paolo Ponzetto.
\newblock 2018b.
\newblock An argument-annotated corpus of scientific publications.
\newblock In {\em Proceedings of the 5th Workshop on Argument Mining}, pages
  40--46, Brussels, Belgium, November. Association for Computational
  Linguistics.

\bibitem[\protect\citename{Levesque \bgroup et al.\egroup
  }2012]{Levesque:2012:WSC:3031843.3031909}
Hector~J. Levesque, Ernest Davis, and Leora Morgenstern.
\newblock 2012.
\newblock The winograd schema challenge.
\newblock In {\em Proceedings of the Thirteenth International Conference on
  Principles of Knowledge Representation and Reasoning}, KR'12, pages 552--561.
  AAAI Press.

\bibitem[\protect\citename{Lin \bgroup et al.\egroup }2014]{pdtbdiscou}
Ziheng Lin, Hwee~Tou Ng, and Min-Yen Kan.
\newblock 2014.
\newblock A pdtb-styled end-to-end discourse parser.
\newblock {\em Natural Language Engineering}, 20(2):151--184.

\bibitem[\protect\citename{Lippi and Torroni}2016]{lippi2016argumentation}
Marco Lippi and Paolo Torroni.
\newblock 2016.
\newblock Argumentation mining: State of the art and emerging trends.
\newblock {\em ACM Transactions on Internet Technology (TOIT)}, 16(2):10.

\bibitem[\protect\citename{Marasovi{\'c} \bgroup et al.\egroup
  }2017]{marasovic-etal-2017-mention}
Ana Marasovi{\'c}, Leo Born, Juri Opitz, and Anette Frank.
\newblock 2017.
\newblock A mention-ranking model for abstract anaphora resolution.
\newblock In {\em Proceedings of the 2017 Conference on Empirical Methods in
  Natural Language Processing}, pages 221--232, Copenhagen, Denmark, September.
  Association for Computational Linguistics.

\bibitem[\protect\citename{Mueller and Thyagarajan}2016]{mueller2016siamese}
Jonas Mueller and Aditya Thyagarajan.
\newblock 2016.
\newblock Siamese recurrent architectures for learning sentence similarity.
\newblock In {\em Thirtieth AAAI Conference on Artificial Intelligence}.

\bibitem[\protect\citename{Opitz and Frank}2018]{opitz-frank-2018-addressing}
Juri Opitz and Anette Frank.
\newblock 2018.
\newblock Addressing the {W}inograd schema challenge as a sequence ranking
  task.
\newblock In {\em Proceedings of the First International Workshop on Language
  Cognition and Computational Models}, pages 41--52, Santa Fe, New Mexico, USA,
  August. Association for Computational Linguistics.

\bibitem[\protect\citename{Opitz and Frank}2019a]{opitz-frank-2019-argument}
Juri Opitz and Anette Frank.
\newblock 2019a.
\newblock An argument-marker model for syntax-agnostic proto-role labeling.
\newblock In {\em Proceedings of the Eighth Joint Conference on Lexical and
  Computational Semantics (*{SEM} 2019)}, pages 224--234, Minneapolis,
  Minnesota, June. Association for Computational Linguistics.

\bibitem[\protect\citename{Opitz and Frank}2019b]{opitzdissec}
Juri Opitz and Anette Frank.
\newblock 2019b.
\newblock Dissecting content and context in argumentative relation analysis.
\newblock In {\em Proceedings of the 6th Workshop on Argument Mining}, pages
  25--34, Florence, Italy, August. Association for Computational Linguistics.

\bibitem[\protect\citename{Peldszus and Stede}2013]{peldszus2013argument}
Andreas Peldszus and Manfred Stede.
\newblock 2013.
\newblock From argument diagrams to argumentation mining in texts: A survey.
\newblock {\em International Journal of Cognitive Informatics and Natural
  Intelligence (IJCINI)}, 7(1):1--31.

\bibitem[\protect\citename{Peldszus and Stede}2015]{peldszus2015joint}
Andreas Peldszus and Manfred Stede.
\newblock 2015.
\newblock Joint prediction in mst-style discourse parsing for argumentation
  mining.
\newblock In {\em Proceedings of the 2015 Conference on Empirical Methods in
  Natural Language Processing}, pages 938--948.

\bibitem[\protect\citename{Peldszus and Stede}2016]{PeldszusStede-ECA:16}
Andreas Peldszus and Manfred Stede.
\newblock 2016.
\newblock An annotated corpus of argumentative microtexts.
\newblock In D.~Mohammed and M.~Lewinski, editors, {\em Argumentation and
  Reasoned Action - Proc. of the 1st European Conference on Argumentation,
  Lisbon, 2015}. College Publications, London.

\bibitem[\protect\citename{Peters \bgroup et al.\egroup
  }2018]{peters-etal-2018-deep}
Matthew Peters, Mark Neumann, Mohit Iyyer, Matt Gardner, Christopher Clark,
  Kenton Lee, and Luke Zettlemoyer.
\newblock 2018.
\newblock Deep contextualized word representations.
\newblock In {\em Proceedings of the 2018 Conference of the North {A}merican
  Chapter of the Association for Computational Linguistics: Human Language
  Technologies, Volume 1 (Long Papers)}, pages 2227--2237, New Orleans,
  Louisiana, June. Association for Computational Linguistics.

\bibitem[\protect\citename{Pollock}1995]{Pollock:1995:CCB:526901}
John~L. Pollock.
\newblock 1995.
\newblock {\em Cognitive Carpentry: A Blueprint for How to Build a Person}.
\newblock MIT Press, Cambridge, MA, USA.

\bibitem[\protect\citename{Skeppstedt \bgroup et al.\egroup
  }2018]{skeppstedt-etal-2018-less}
Maria Skeppstedt, Andreas Peldszus, and Manfred Stede.
\newblock 2018.
\newblock More or less controlled elicitation of argumentative text: Enlarging
  a microtext corpus via crowdsourcing.
\newblock In {\em Proceedings of the 5th Workshop on Argument Mining}, pages
  155--163, Brussels, Belgium, November. Association for Computational
  Linguistics.

\bibitem[\protect\citename{Socher \bgroup et al.\egroup
  }2013]{socher-etal-2013-recursive}
Richard Socher, Alex Perelygin, Jean Wu, Jason Chuang, Christopher~D. Manning,
  Andrew Ng, and Christopher Potts.
\newblock 2013.
\newblock Recursive deep models for semantic compositionality over a sentiment
  treebank.
\newblock In {\em Proceedings of the 2013 Conference on Empirical Methods in
  Natural Language Processing}, pages 1631--1642, Seattle, Washington, USA,
  October. Association for Computational Linguistics.

\bibitem[\protect\citename{Stab and Gurevych}2014]{stab2014identifying}
Christian Stab and Iryna Gurevych.
\newblock 2014.
\newblock Identifying argumentative discourse structures in persuasive essays.
\newblock In {\em Proceedings of the 2014 Conference on Empirical Methods in
  Natural Language Processing (EMNLP)}, pages 46--56.

\bibitem[\protect\citename{Stab and Gurevych}2017]{stab2017parsing}
Christian Stab and Iryna Gurevych.
\newblock 2017.
\newblock Parsing argumentation structures in persuasive essays.
\newblock {\em Computational Linguistics}, 43(3):619--659.

\bibitem[\protect\citename{Swanson \bgroup et al.\egroup
  }2015]{swanson2015argument}
Reid Swanson, Brian Ecker, and Marilyn Walker.
\newblock 2015.
\newblock Argument mining: Extracting arguments from online dialogue.
\newblock In {\em Proceedings of the 16th annual meeting of the special
  interest group on discourse and dialogue}, pages 217--226.

\bibitem[\protect\citename{Vaswani \bgroup et al.\egroup }2017]{NIPS2017_7181}
Ashish Vaswani, Noam Shazeer, Niki Parmar, Jakob Uszkoreit, Llion Jones,
  Aidan~N Gomez, \L~ukasz Kaiser, and Illia Polosukhin.
\newblock 2017.
\newblock Attention is all you need.
\newblock In I.~Guyon, U.~V. Luxburg, S.~Bengio, H.~Wallach, R.~Fergus,
  S.~Vishwanathan, and R.~Garnett, editors, {\em Advances in Neural Information
  Processing Systems 30}, pages 5998--6008. Curran Associates, Inc.

\bibitem[\protect\citename{Wachsmuth \bgroup et al.\egroup
  }2018]{wachsmuth-etal-2018-argumentation}
Henning Wachsmuth, Manfred Stede, Roxanne El~Baff, Khalid Al-Khatib, Maria
  Skeppstedt, and Benno Stein.
\newblock 2018.
\newblock Argumentation synthesis following rhetorical strategies.
\newblock In {\em Proceedings of the 27th International Conference on
  Computational Linguistics}, pages 3753--3765, Santa Fe, New Mexico, USA,
  August. Association for Computational Linguistics.

\bibitem[\protect\citename{Walton}2009]{wd2009}
Douglas Walton.
\newblock 2009.
\newblock Objections, rebuttals and refutations.
\newblock In {\em Argument Cultures: Proceedings of the 8th OSSA Conference},
  pages 1--10, Windsor, Ontario.

\end{thebibliography}

\end{document}